\documentclass[letterpaper, 10 pt, conference]{ieeeconf}  % Comment this line out if you need a4paper

\usepackage{amsmath}
\usepackage{amsfonts}
\usepackage{siunitx}
\usepackage{subcaption}
\usepackage{subfig}
\usepackage{graphicx}
\usepackage{makecell}
\usepackage{url}
\usepackage{multicol}
\usepackage{multirow}
\usepackage{adjustbox}

\IEEEoverridecommandlockouts                              % This command is only needed if 
\title{\LARGE \bf
KYON: Semi-Modular Wheel–Legged Quadruped With Agile Bi-Manual Capability   %or "KYON: Agile Wheel–Legged Quadruped with Bi-Manual Dexterity //and Reconfigurable Legs"
}

\author{Luca Rossini$^{1}$, Arturo Laurenzi$^{1}$, Francesco Ruscelli$^{1}$, Yifang Zhang$^{1}$, Jingcheng Jiang$^{1}$, Giovanbattista Gravina$^{1}$, \\ Lorenzo Baccelliere$^{1}$, Corrado Burchielli$^{1}$, Stefano Cordasco$^{2}$, Luca Muratore$^{1}$, and Nikos Tsagarakis$^{1}$% <-this % stops a space
\thanks{$^{1}$Humanoids and Human Centered Mechatronics, Fondazione Istituto Italiano di Tecnologia, Via Morego 30, 16163, Genova, Italy}
\thanks{${^2}$ Advanced Robotics Facility, Fondazione Istituto Italiano di Tecnologia, Via Morego 30, 16163, Genova, Italy}
\thanks{\tt\small \{name.surname\}@iit.it}}%
 
\begin{document}

\maketitle
\thispagestyle{empty}
\pagestyle{empty}

%%%%%%%%%%%%%%%%%%%%%%%%%%%%%%%%%%%%%%%%%%%%%%%%%%%%%%%%%%%%%%%%%%%%%%%%%%%%%%%%
\begin{abstract}
This paper presents KYON, a hybrid wheel–legged quadruped robot equipped with a bi-manual upper body for loco-manipulation tasks. The platform features a semi-modular design with reconfigurable lower legs, enabling both wheeled and legged locomotion depending on the environment. A design approach that places actuators in the base and uses transmission mechanisms reduces distal inertia, improving agility and dynamic performance.
The robot integrates a whole-body control framework together with a reinforcement learning-based policy to handle nonlinear dynamics and enhance robustness to disturbances for the execution of locomotion and manipulation tasks, independently. Experimental results demonstrate effective dynamic locomotion and bi-manual manipulation, validating the platform’s capability to operate in complex and unstructured scenarios.
%The code material to simulate and control KYON will be released after acceptance to comply with the double blind review policy.
\end{abstract}

%%%%%%%%%%%%%%%%%%%%%%%%%%%%%%%%%%%%%%%%%%%%%%%%%%%%%%%%%%%%%%%%%%%%%%%%%%%%%%%%
\section{INTRODUCTION}
\label{sec:introduction}
Legged robots have the potential to make a significant impact in industrial, domestic, and natural environments, all of which are inherently designed by and for humans. 
Although their under-actuation introduces several challenges, humanoids and quadrupeds can exploit human-inspired embodiment to navigate steps and perform multi-contact loco-manipulation strategies that humans use daily. 
Recent advancements have demonstrated the deployment of legged robots in various scenarios, highlighting both their applicability and growing market interest~\cite{spot, anymal-x, b2}.
\begin{figure}
\begin{subfigure}{\columnwidth}
    \centering
    \includegraphics[width=\columnwidth]{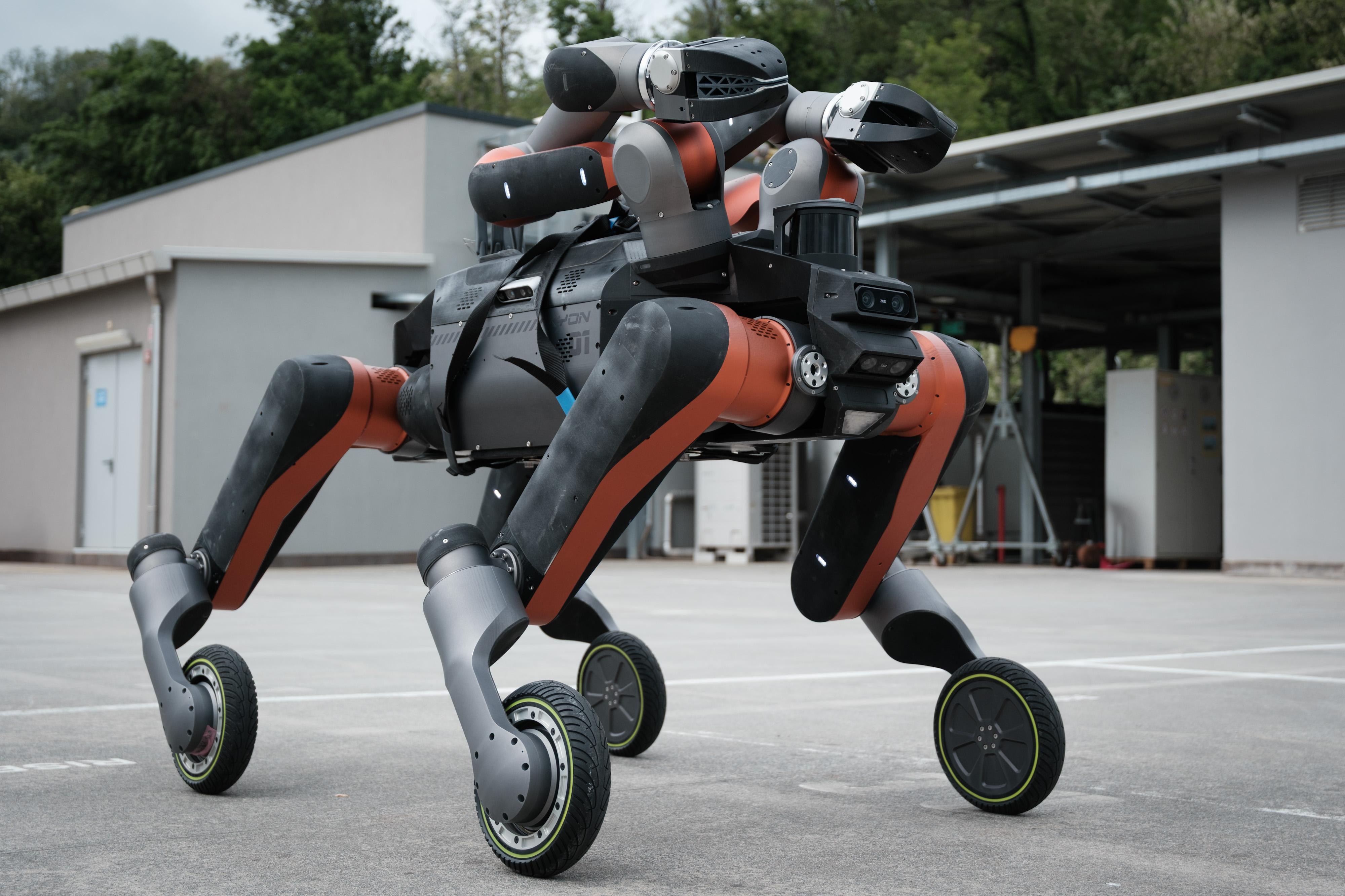}
    \caption{\label{subfig:kyon_wheel}}
    \label{subfig:kyon_wheel}
\end{subfigure}
\\
\begin{subfigure}{\columnwidth}
    \centering
    \includegraphics[width=\columnwidth]{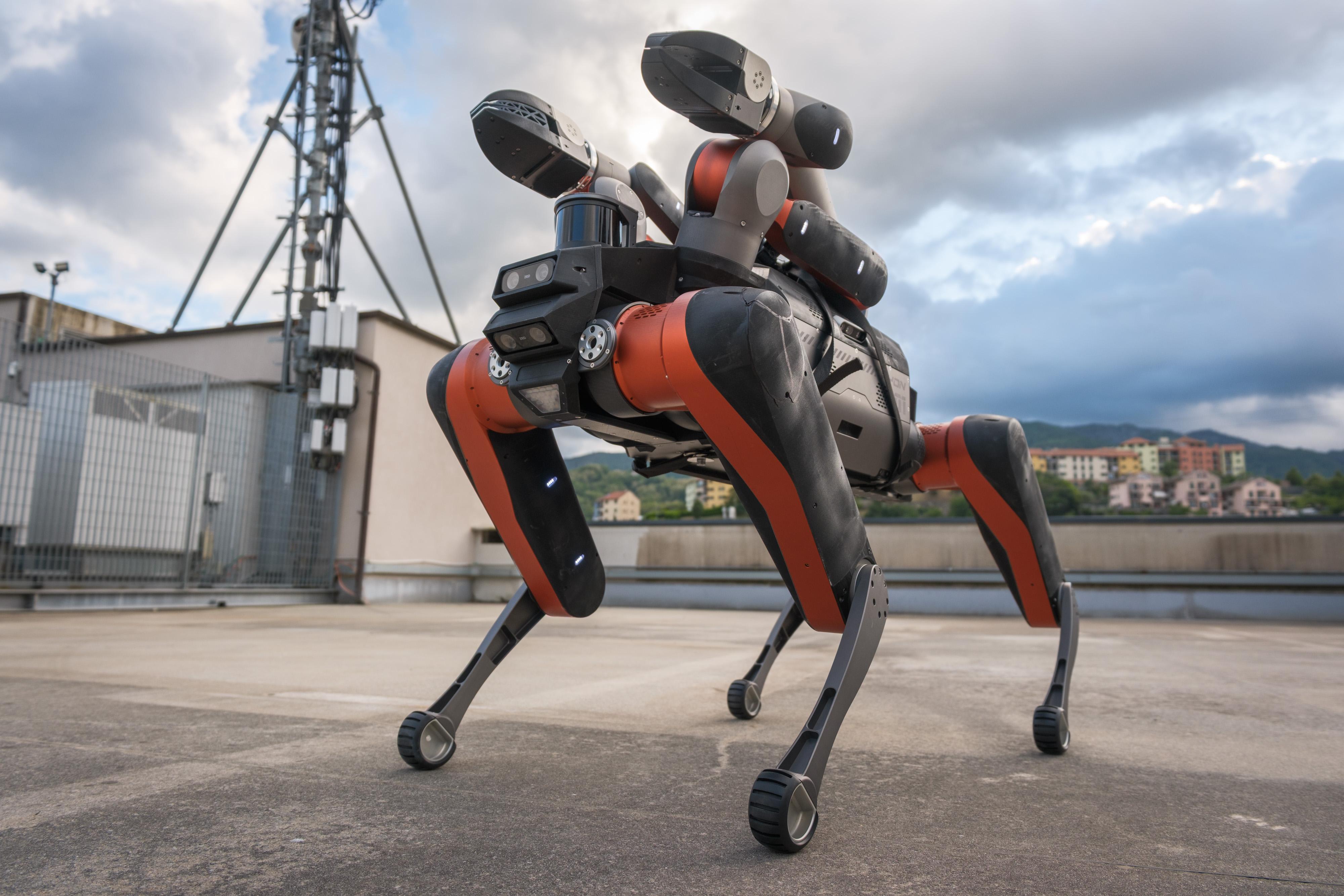}
    \caption{}
    \label{subfig:kyon_leg}
\end{subfigure}
\caption{The KYON robot in \ref{subfig:kyon_wheel} wheeled and \ref{subfig:kyon_leg} legged configuration.}
\label{fig:kyon}
\end{figure}
\par A major limitation of current legged systems lies in their limited payload capacity, often compromised in favor of dynamic performance. 
To address this issue, hybrid wheel–legged platforms have been proposed as an effective solution for reducing locomotion energy costs~\cite{kashiri:centauro, valsecchi:barry}. 
Wheeled locomotion requires only a subset of actuators and avoids the need for continuous whole-body coordination to maintain balance, while legged locomotion remains essential to achieve dynamic mobility in human-centered environments.
Yet, quadrupeds are still far from matching human-like performance in heavy-payload task, an essential requirement to substitute human operator in stressful scenarios, thus reducing the occurrence of injuries and accidents.
Furthermore, most of the quadruped embodiments comprehend a dynamic legged lower-body and has never been provided with bi-manual capabilities able to mimic the human characteristics.
\par The rest of this paper presents the current SoA about high-payload quadruped robots in Sec.\ref{sec:soa}.
Then, the motor sizing for both the lower and upper-body actuators is described in Sec.\ref{sec:lb_codesign} and Sec.\ref{sec:ub_codesign}, respectively. 
Sec.\ref{sec:mechatronics} and Sec.\ref{sec:control} describe the mechatronics and control framework.
Finally, Sec.\ref{sec:experiments} presents a list of validative experiments to assess the final performance of the robot dealing with manipulation and locomotion tasks, and Sec.\ref{sec:conclusions} summarizes the outcome of this letter.
\section{PREVIOUS WORK}
\label{sec:soa}
\begin{figure}
    \centering
    \includegraphics[width=\columnwidth]{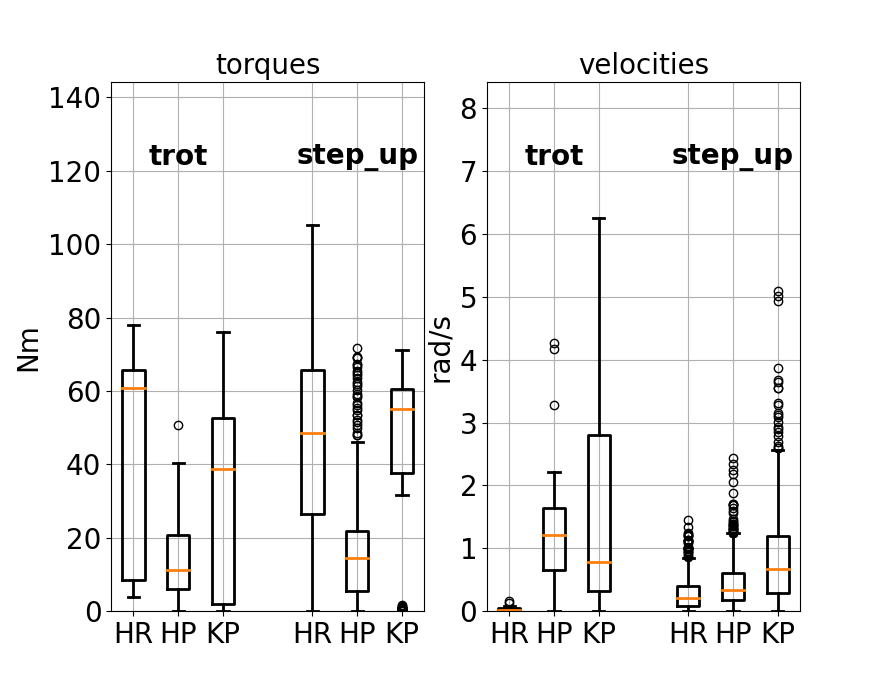}
    \caption{Link-side torques and velocities boxplots generated while trotting and stepping up.}
    \label{fig:lb_torque_vel_pow}
\end{figure}
\begin{table*}
\centering
% \begin{adjustbox}{width=\columnwidth}
    \begin{tabular}{ | c | c | c c | c c | c c | c c |}
    \hline
    \multirow{2}{*}{MOTOR} & \multirow{2}{*}{$P_N$ (\si{\watt})} & \multicolumn{2}{c|}{$\tau_c$ (\si{\newton\meter})} & \multicolumn{2}{c|}{$\tau_p$ (\si{\newton\meter})} & \multicolumn{2}{c|}{$\omega_{\mathrm{max}}$ (\si{\radian\per\second})} & \multicolumn{2}{c|}{M (\si{\kilo\gram)}} \\
    & & 1:30 & 1:50 & 1:30 & 1:50 & 1:30 & 1:50 & 1:30 & 1:50\\
    \hline
    TBM-2G-08513D & 734 & 37.1 & 55 & 99.9 & 166.5 & 16.2 & 9.7 & 0.94 & 0.64 \\
    RI100 T-Motor & 720 & 39.6 & 55 & 111 & 185 & 14.3 & 8.6 & 1.04 & 0.74 \\
    RI80 T-Motor  & 492 & 39.6 & 55 & 111 & 185 & 8.7 & 5.2 & 1.04 & 0.74 \\
    TMCustom 8116 & 686 & 41.6 & 55  & 123.7 & 186 & 9.8 & 5.9 & 0.86 & 0.56 \\
    TMCustom 8120 & 1089 & 60.7 & 55 & 168.7 & 186 & 9.9 & 5.94 & 0.96 & 0.66 \\
    \hline
    \end{tabular}
% \end{adjustbox}
\caption{Motors and drive combinations selected for the lower-body design. Maximum torques are computed considering both motor and drive limits. The drives considered are the CPL series size 25 Harmonic drive ($N=1:50$, $\tau_c = $\qty{55}{\newton\meter}, $\tau_p = $\qty{186}{\newton\meter}, $m = $\qty{0.24}{\kilo\gram}) and the CPL series 32 Harmonic drive ($N=1:30$, $\tau_c = $\qty{75}{\newton\meter}, $\tau_p = $\qty{200}{\newton\meter}, $m = $\qty{0.54}{\kilo\gram}). Motor velocity limits are considered at the nominal voltage of \qty{48}{\volt}.}
\label{tab:motors}
\end{table*}
Wheeled and tracked robots, although characterized by fast locomotion and high payload capacity on flat surfaces, experience a significant drop in performance when operating on soft or uneven terrain.
Legged robots have been regarded as a natural evolution, expanding robotic workspaces beyond confined and structured environments and enabling traversal across a wide variety of terrains.
\mbox{Boston Dynamics}' BigDog~\cite{raibert:bigdog}, WildCat~\cite{wildcat}, and LS3~\cite{ls3} represent a clear milestone in this direction, as the first quadrupedal platforms capable of carrying substantial payloads in outdoor environments.
These achievements were made possible by high power-density Internal Combustion Engines (ICEs) combined with hydraulic actuation. However, the exhaust emissions produced by ICEs limit their use in indoor environments.
The integration of electric motors with hydraulic actuation preserved high power density while eliminating the drawbacks associated with ICE-based power generation. Over the years, several hydraulically actuated robots have been developed.
Notable examples include the HyQ series~\cite{semini:hyd_leg, semini:hyq, semini:hyq2max}, developed at the Istituto Italiano di Tecnologia (IIT), which demonstrated remarkable power performance by pulling a stationary \qty{3300}{\kilo\gram} airplane.
Similarly, SCalf~\cite{scalf} and Big Elephant~\cite{big_elephant}, developed by Shandong University and Shanghai Jiao Tong University, respectively, were among the first robots to demonstrate fast trotting over challenging terrain and were also the first rideable quadrupeds.
Operating hydraulically actuated robots requires regulating fluid flow, monitoring pressure levels, coordinating valve operation, and adjusting flow rates and pressure based on feedback signals to achieve the desired motion.
Nevertheless, hydraulic drive systems are typically bulky, heavy, and noisy, which limits their widespread adoption.
\par Recent advances in electric actuation have made electrically powered quadruped robots suitable for commercial applications, particularly in inspection and entertainment.
Examples include \mbox{Boston Dynamics}' Spot~\cite{spot}, ANYbotics' ANYmal~\cite{anymal-x}, and DEEP Robotics' Lite, Lynx, and X Series~\cite{deep_robotics}.
Compared to hydraulically actuated quadrupeds, electrically powered systems offer advantages such as a more compact design, lower noise levels, and simplified control.
However, their reduced payload-to-weight ratio limits their deployment in demanding scenarios requiring advanced manipulation capabilities.
The limited manipulation capabilities of quadruped robots have been addressed from both hardware and software perspectives.
On the hardware side, dedicated robotic arms can be mounted on quadrupeds to enable manipulation~\cite{fu:quadruped_manipulation, mittal:quadruped_manipulation}. However, this approach increases the overall mass and reduces the available payload capacity. As a result, most existing solutions incorporate a single arm, which restricts dexterous performance compared to human capabilities.
To improve payload efficiency and locomotion performance, wheeled–quadrupedal manipulators have been introduced, reducing transportation costs while enabling fast and stable mobility~\cite{jiang:wheeld_quadrupedal_manipulator}.
Nevertheless, legged locomotion remains essential in highly cluttered environments where wheels may fail to establish reliable contact (e.g., climbing small steps or steep stepping stones).
\par From a software perspective, approaches such as~\cite{pedipulation, feng:pushing_whole_body} have proposed control strategies that allow interaction with objects using a leg or the whole body, without requiring external manipulators.
However, these methods are limited to non-prehensile manipulation tasks, such as pushing and reorienting objects.
\par In this work, we introduce KYON, a novel hybrid wheel-legged quadruped robot equipped with a bi-manual manipulation system and a steering-wheel mechanism. In particular, KYON features the following key design elements:
\begin{itemize}
\item Semi-modular body design
\item Reconfigurable lower legs
\item Capability for both external and internal knee configurations
\item Omni-directional wheeled mobility
\item Bi-manual manipulation
\end{itemize}
The distal wheeled modules are designed for straightforward replacement with standard passive legs, enabling more dynamic motions by reducing the inertial properties of each limb for highly cluttered scenarios.
With the wheeled configuration mounted, the robot has a total mass of \qty{90}{\kilo\gram} and a standing height of \qty{0.8}{\meter}.
Both the legs and arms are actuated with five DoFs each.
The end-effectors are custom single-DoF, torque-controllable grippers.
To the best of our knowledge, KYON is the first hybrid wheel-legged quadruped robot integrating a dedicated bi-manual system for high-performance manipulation.
%
% \subsection{System Requirements}
% \label{subsec:sys_specs}
% KYON has been designed following well-defined specifications in terms of locomotion and manipulation capabilities, and payload capacity. Specifically, the robot has to:
% %
% \begin{itemize}
%     \item Trot at a minimum speed of \qty{1}{\kilo\meter\per\hour}.
%     \item Carry a total payload of \qty{28}{\kilo\gram} divided in \qty{4}{\kilo\gram} per each arm, and \qty{20}{\kilo\gram} carried on the robot base.
%     \item Cross \qty{0.2}{\meter} wide gaps.
%     \item Step up a \qty{0.3}{\meter} high obstacle.
% \end{itemize}
%
\section{LOWER-BODY MOTOR SIZING}
\label{sec:lb_codesign}
The main objective of the lower-body motor sizing process was to meet the system-level requirements while respecting the imposed kinematic constraints. 
%The robot was designed to reach a height of \qty{0.8}{\meter} from the ground when fully extended. This configuration results in both the thigh and calf links being \qty{0.35}{\meter} long, with a wheel radius of approximately \qty{10}{\centi\meter}.
%
\par For actuator selection, the platform is required to trot at a nominal velocity of \qty{1.0}{\meter\per\second} and to climb a \qty{0.3}{\meter} step while carrying a nominal payload of \qty{30}{\kilo\gram} mounted on the base. The analysis started from a URDF model that included the kinematic structure together with an initial estimate of the mass distribution. This model was employed within the Optimal Control (OC) framework Horizon~\cite{ruscelli:horizon} to compute offline trajectories for the two locomotion tasks described above.
The payload was represented in simulation by adding an additional URDF link with appropriate mass and inertia properties, rigidly attached to the base. The resulting trajectories were then evaluated in terms of link-side torques and velocities, which served as the primary metrics for motor and gearbox selection. Fig.~\ref{fig:lb_torque_vel_pow} reports the link-side torque and velocity profiles, together with the mechanical power required during the trotting and step-climbing maneuvers.
\par Based on this initial analysis, we pre-selected a set of motors whose nominal mechanical output exceeds the power required for the locomotion tasks, which is approximately \qty{150}{\watt}.
To account for electrical and mechanical losses and to adopt a more conservative selection criterion, we further restricted the candidates to motors with a nominal mechanical output above \qty{200}{\watt}. The lower-body simulations indicate a required continuous torque ranging between \qty{50} and \qty{60}{\newton\meter}, with peaks exceeding \qty{100}{\newton\meter}.
Under these conditions, a $1:30$ reduction ratio is generally not feasible, with the exception of the TMCustom 8120. In contrast, motors paired with a reduction ratio of $N=1:50$ satisfy the torque requirements with a comfortable margin. However, among these options, only the TBM-2G-08513D and the RI100 provide sufficiently high output velocity.
Among the available combinations, we selected the solution offering the highest maximum torque. This choice enables higher acceleration capability and provides additional margin for carrying heavier payloads, particularly considering the mass of the two arms and the steering wheel modules.
\section{UPPER-BODY CODESIGN}
\label{sec:ub_codesign}
\begin{figure}
    \centering
    \includegraphics[width=\columnwidth]{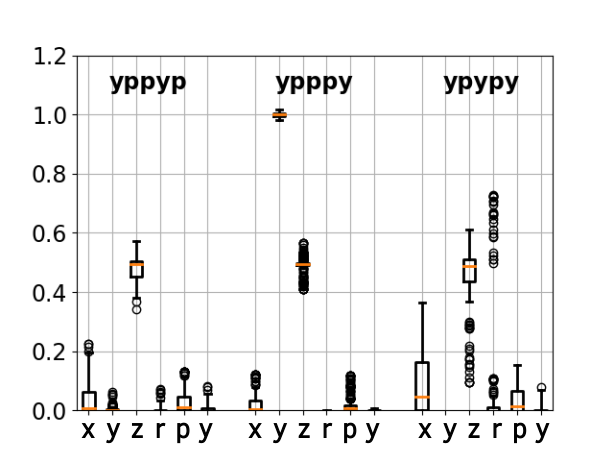}
    \caption{Box-plot of the base motion along the six-axes for the three arms' configurations.}
    \label{fig:fb_motion}
\end{figure}
%
% \begin{figure}
%     \centering
%     \includegraphics[width=0.8\columnwidth]{figures/ky_upper-body 1.pdf}
%     \caption{Graphical representation of the KYON's upper body. Yaw and pitch joints are represented with blue and green arrows respectively.}
%     \label{fig:kyon_upper_body}
% \end{figure}
%
To meet the upper-body mass constraint, which is limited to \qty{18}{\kilo\gram}, we opted for a pair of 5-DoF arms.
Each arm was designed with a total reach of \qty{0.9}{\meter}, ensuring a sufficient workspace to perform the necessary manipulation tasks.
The primary challenge of using an underactuated configuration—due to the omission of one joint—was identifying a joint layout that avoids kinematic singularities and maintains good manipulability in the frontal workspace.
By limiting motion at the base, we reduce the torque demands on the legs, decreasing the likelihood of instability or falls during manipulation.
To address this, we evaluated three potential arm configurations by observing base movement as the end-effector executed standard grasping motions within the frontal workspace.
All three configurations shared a common structure: a yaw–pitch shoulder and a pitch elbow.
The first design (YPYPY) featured an anthropomorphic shoulder, an additional proximal yaw joint, and a distal yaw joint acting as a wrist.
The second and third configurations introduced a 2-DoF wrist to enhance end-effector control.
One used a pitch–yaw wrist (YPPPY), while the other reversed the wrist joint sequence (YPPYP).
For control, the robot utilized CartesI/O~\cite{cartesio}, a QP-based Inverse Kinematics (IK) solver.
Cartesian trajectories were generated by manipulating an interactive marker in rViz, exploring a variety of end-effector poses for each candidate configuration.
The corresponding base motion across six degrees of freedom is illustrated in Fig.~\ref{fig:fb_motion}.
%
% \begin{figure}
% \centering
% \includegraphics[width=\columnwidth]{figures/random_poses.png}
% \caption{Link-side torques from the random arm motion while handling a payload of \qty{4}{\kilo\gram}}
% \label{fig:upper_body_torques}
% \end{figure}
%
Tests showed that the YPPPY configuration resulted in the least base displacement during left-arm manipulation.
To increase the overall reachable workspace and promote the engagement of pitch joints during bimanual tasks, the two arms were mounted with a \qty{20}{\degree} tilt relative to the vertical axis.
\begin{table}
\centering
\begin{tabular}{ | c | c | c | c  c | c | c |}
\hline
\multirow{2}{*}{MOTOR} & \multirow{2}{*}{Joint} & \multirow{2}{*}{$N$} & $\tau_c$  & $\tau_p$ & $\omega_{\mathrm{max}}$ & $M$  \\
& & & \multicolumn{2}{c|}{(\si{\newton\meter})} & (\si{\radian\per\second}) & (\si{\kilo\gram})\\
\hline
\multirow{2}{*}{RI60 T-Motor} & 2 & 1:120 & 49 & 146.7 & 3.9 & 0.3 \\
 & 1, 3 & 1:100 & 39 & 110 & 4.7 & 0.27 \\
RBE-1210A & 4, 5& 1:100 & 7.7  & 25.6 & 4.7 & 0.18 \\
\hline
\end{tabular}
\caption{Motors and drive combinations selected for the upper-body design. Maximum torques are computed considering both motor and drive limits. The drives considered are the CPL series size 20 Harmonic drive ($N=1:120$, $\tau_c = $ \qty{49}{\newton\meter}, $\tau_p = $ \qty{147}{\newton\meter}, $m = $ \qty{0.14}{\kilo\gram}) and the CPL series 17 Harmonic drive ($N=1:100$, $\tau_c = $ \qty{39}{\newton\meter}, $\tau_p = $ \qty{110}{\newton\meter}, $m = $ \qty{0.1}{\kilo\gram}), and the CSD series 14 harmonic drive ($N=1:100$, $\tau_c = $ \qty{7.7}{\newton\meter}, $\tau_p = $ \qty{31}{\newton\meter}, $m = $ \qty{0.06}{\kilo\gram}). Motor velocity limits are considered at the nominal voltage of \qty{48}{\volt}. Joint numbering is in Fig.~\ref{fig:upperbody_overview}.}
\label{tab:ub_motors}
\end{table}
To meet the resulting torque demands, the selected motor–drive combinations are listed in Tab.~\ref{tab:ub_motors}.
\section{MECHATRONICS}
\label{sec:mechatronics}
\subsection{Leg subsystem}
\begin{figure}
    \centering
    \includegraphics[width=\columnwidth]{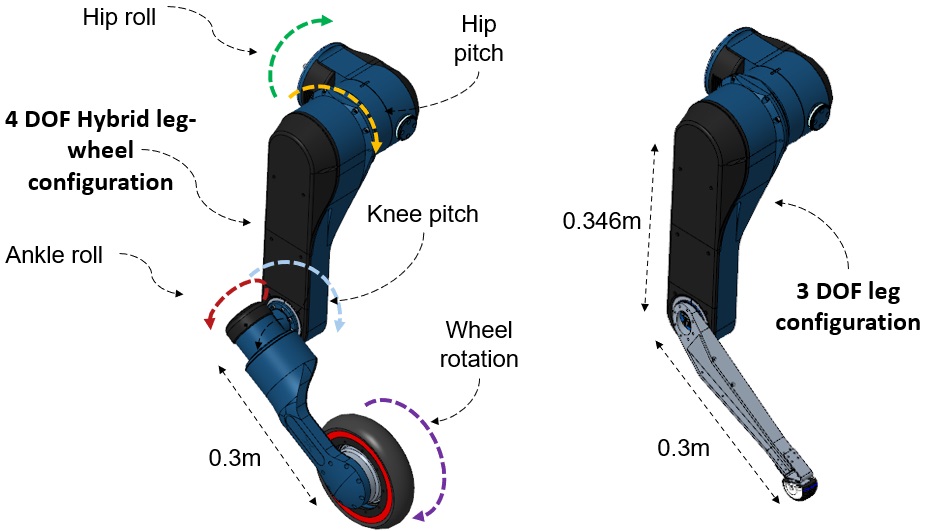}
    \caption{The two configurations of the leg showing the the wheel and foot type of end-effector employed in the two variants.}
    \label{fig:legconfigurations}
\end{figure}
Following the study of the existing four legs robotic platforms performed during the first months of the project the under development platform will make use of symmetrical legs in which the upper and the lower leg links have approximately the same length.
In particular, the length of the upper leg was set to \qty{0.346}{\meter} while the length of the lower leg from the knee joint to the interface of the foot or wheel end-effector was set to \qty{0.3}{\meter}. This results in an overall leg length of \qty{0.646}{\meter} without the wheel or the foot end-effector. Considering the wheel module the overall length of the leg including the wheel radius is \qty{0.746}{\meter}. 
The overall mechanical assembly of the leg subsystem is introduced in Fig.~\ref{fig:legconfigurations}.
\begin{figure}
    \centering
    \includegraphics[width=\columnwidth]{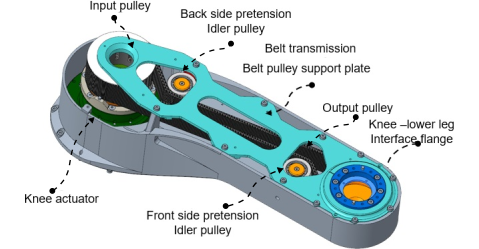}
    \caption{The front side of the upper leg assembly showing the details of the belt transmission system and its components. }
    \label{fig:upperlegfront}
\end{figure}
\par The upper leg module is common between the two leg configurations. It implements the main
part of the leg kinematics that includes a hip complex with two degrees of freedom (DOFs) and a
knee joint providing an additional degree of freedom. All the three joints of the upper leg are
powered by the same type of actuator described earlier.
\par The knee actuator torque and power is delivered to the knee joint through a secondary high power belt transmission system integrated in the output of the knee actuator (after the primary harmonic reduction drive), in Figure \ref{fig:upperlegfront}. To enable adequate tuning range of the pretension of the belt transmission two idler pretension pulley are considered, one placed at the back side (closely to the knee actuator) and one placed in the front side closely to the knee pulley flange. The two idler pretension pulleys are custom designed to reduce overall size and mass. The pretension mechanism is realized with the use of a fine thread screw that acts in a prismatic guide to constrain and control the position of the idler pulley with the respect to the belt. A belt pulley support plate has been considered and integrated to provide double support of the input (knee actuator side at the hip level) and output pulley (at the knee level) of the belt reducing the axial loading of the related bearings and increasing the stiffness of the overall upper leg assembly. 
\par
\begin{figure}
    \centering
    \includegraphics[width=\columnwidth]{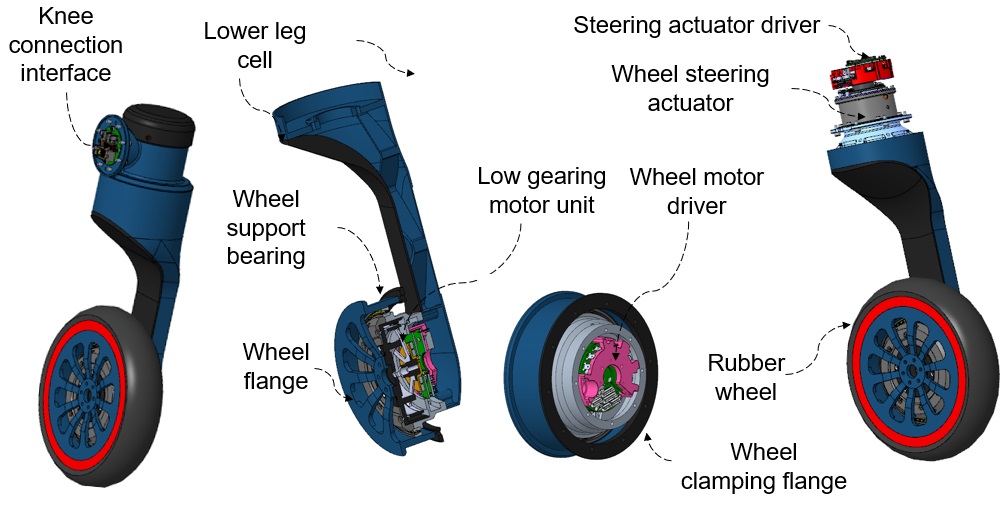}
    \caption{The lower leg wheel module showing the connection interface and the steering-wheel joints actuation. }
    \label{fig:lowerleg_wheel}
\end{figure}

 As described earlier, the lower leg assembly will be produced in two configurations.
The first configuration is composed by the lower leg cell structure, the wheel end-effector module and the wheel steering joint, Figure \ref{fig:lowerleg_wheel}. 
The wheel steering joint is located at the level of the knee enabling the rotation of the lower leg cell and wheel along an axis that intersects with the wheel rotation axis. The steering joint is powered by the same actuation unit that is used in shoulder yaw and elbow joints of the arm. The lower leg cell in this configuration is connected to the output of the steering actuator and terminates to the interface where the wheel actuator is fixed. 
A tubeless tire is mounted on the output of the wheel actuator through the wheel flange interface. The tubeless rubber tire is fixed on the wheel flange using the clamping flange, Figure \ref{fig:lowerleg_wheel}. The wheel tire diameter is \qty{0.2}{\meter} trading off between wheel weight and capability to overcome terrain unevenness. 
The actuation of the wheel module is provided by a modified commercial actuator from TMotor company.  In particular the actuation drive model AK80-8 was customized in terms of housing in order to be integrated inside the lower leg cell. The customization in the housing of the commercial drive targeted to realize a modular actuation- wheel module assembly while integrating the actuation drive in the lower leg cell. This enables a lightweight implementation that is critical for keeping low the weight of the wheel module while at the same time it enables easy exchange of the wheel tire component, e.g. for replacing with a larger diameter wheel.
\par The second configuration comes with a passive lower leg link with a circular extremity to guarantee point contact at every angular position of the knee joint.
\subsection{Arm subsystem}
\begin{figure}
    \centering
    \includegraphics[width=\columnwidth]{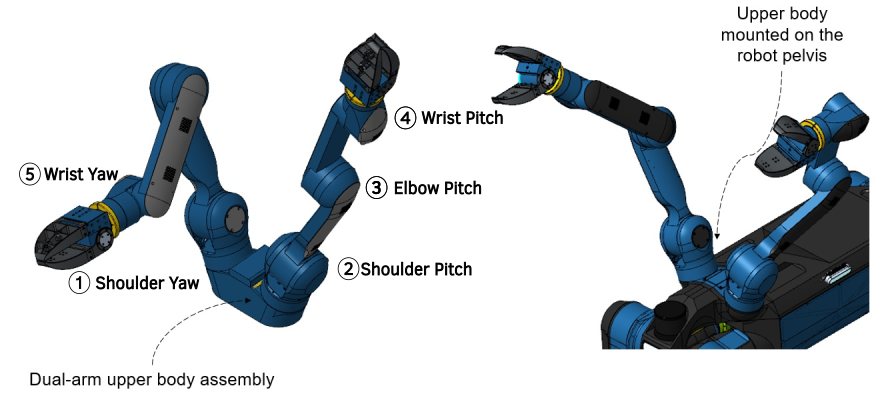}
    \caption{Overview of the dual arm upper body subsystem. }
    \label{fig:upperbody_overview}
\end{figure}
The arm module is composed of three distinct body sections presenting the upper arm, the lower arm and the wrist link sections. The length of the upper arm, lower arm and wrist sections is \qty{0.32}{\meter},  \qty{0.25}{\meter},    \qty{0.13}{\meter} respectively resulting it a total arm length of \qty{0.7}{\meter}.
Finally, the length of the end-effector module is \qty{0.2}{\meter} resulting in an overall length of the arm of \qty{0.9}{\meter} when the end-effector is also included.
The arms' kinematics follows the joint scheme described in the previous section (Fig.~\ref{fig:upperbody_overview}.
to extend the range of motion of the elbow joint, the upper arm and lower arm links are implemented to move on different planes permitting to significantly extend the range of motion of the elbow joint. This extended range of motion enables a compact stowed  configurations for the arm module at the back of the robot pelvis when the arm module is not used.
Furthermore, the range of motion for the joints of the arm were maximized at the extend possible during the design study to facilitate the manipulation motion range of the arm module. The shoulder yaw provides a range motion of $[-160,+160]\, deg$. The range of motion of the shoulder pitch joint is $[-125,+125]\, deg$ while the elbow joint can move within the range of  $[-160,+160]\, deg$.  Finally, the range of the wrist joints are $[-80,+80]\, deg$ and $[-160,+160]\, deg$ for the pitch and the roll joint respectively.
\par For the interconnection of the actuator units, the arm design followed an exoskeleton link structure approach.  Exoskeleton link cells have higher structural stiffness and minimize the effect of unmeasured elasticity. In the principle of the exoskeleton link structure and actuation integration, the body of the actuators are floating inside this exoskeleton cell structure while the actuation drives are fixed to the structures of the previous and next link.  In particular, the actuation modules are mechanically interfaced to the exoskeleton link cell structures using two flange interfaces in the body of the actuation drive, one fixed to the previous link and the other to the subsequent link. 
\par The design of the shoulder integrates two joints, the shoulder yaw and the shoulder pitch joint. The body of the actuator is mounted on the upper body mounting structure while the output flange of the shoulder yaw drive is fixed to the structure cell that embraces the shoulder pitch joint. The motor driver of the shoulder yaw drive is placed inside the upper body mounting structure. 
A fork cell structure that provides double support on the shoulder pitch joint is employed to enhance structural stiffness in this proximal joint of the robot arm. 
The output flange of the shoulder pitch actuation drive is fixed to one side of the fork cell while the body of the actuation drive is fixed to the upper arm link cell.
As a result, the body shoulder pitch actuation drive is rotating together with the upper arm link cell. 
The choice of this mounting was made to facilitate the wiring between the shoulder pitch actuation drive and its motor driver electronics, which are fixed in the upper arm link cell structure.

\par The upper arm link cell structure therefore integrates the body of the shoulder pitch actuation drive at one end, and the elbow pitch actuation drive at the distal end.
The elbow joint links the upper arm section with the lower arm section, whose actuator is fixed in the upper arm link cell structure while the output flange of the elbow actuation drive is connected to the lower arm link cell structure. The lower arm link cell structure integrates the body of the wrist pitch actuation drive at its distal end. The motor driver of the wrist pitch actuator is located inside the lower arm link structure facilitating the wiring from the actuator to the motor driver electronics.
\par Finally, the wrist link represents the last section of the arm subsystem. The wrist link cell structure is connected to the output of the wrist pitch actuation drive and houses the wrist roll actuator that drives the last joint of the arm. 
The motor driver of the last joint is located inside the wrist link cell structure at the height of the wrist pitch connection flange. 
In the output of the wrist roll actuator a wrist flange providing a modular interconnection interface permits the quick mounting and removal of the end-effector.
\par The end-effector module of the arm is based on a custom modular gripper design composed of a main gripper module and a pair of modular jaws. 
In this modular jaw-type gripper the bottom jaw is fixed while the top jaw is powered by a single actuator. Both the bottom and top jaw have been designed and integrated into the main module of the gripper in a modular/re-configurable manner, permitting to interchange or replace jaws with
others of different geometry if required to adapt the jaw geometry to the shape and size of a particular object to grasp or for providing different mechanical properties in the jaws, e.g. by fabricating the jaws from different
softer material.
\section{SOFTWARE AND CONTROL ARCHITECTURE}
\label{sec:control}
Kyon's software architecture is based on a combination of model-based and learning algorithm to manage locomotion and manipulation tasks independently.
% Compared to our previous works on closed loop Model Predictive Control (MPC)~\cite{rossini:talos}, we also
For legged locomotion purposes, we developed a high-level controller based on a Deep Reinforcement Learning policy to handle unmodeled events coming from the environment.
The high-level control policy is trained using the Mujoco XLA (MJX) framework randomizing the delay of each actuator to simulate the communication delays and the safety low-pass filter of our middle-level controller.
Additionally, domain randomization is performed on each body and actuator of the robot articulation, including mass, inertia, friction, and actuator's gains. 
\par The policy takes in input a planar velocity command vector $\mathbf{\hat{v}_b} \in SE(2)$ and it is trained using an asymmetric actor-critic agent. The actor inputs noisy proprioceptive information $\mathbf{o}_t \in \mathbb{R}$ including the base angular velocity $^B\mathbf{\omega}_b \in \mathbb{R}^3$, the projected gravity vector $\mathbf{g} \in \mathbb{R}^3$, the joint positions relative to the initial one $\mathbf{q} \in \mathbb{R}^{12}$, the joint position error of the last three steps $(\mathbf{q} - \mathbf{\hat{q}}) \in \mathbb{R}^{3\cdot12}$, the position of each of the $i$-th foot in base frame $^B\mathbf{p}_{c,i} \in \mathbb{R}^{4\cdot3}$, the velocity command $\mathbf{\hat{v}_B} \in SE(2)$, the phase vector to track the desired gait frequency as in~\cite{lee:learning} $\mathbf{\varphi}_i \in \mathbb{R}^{4\cdot3}$, and the previous action executed $\mathbf{a}_{t-1} \in \mathbb{R}^{12}$.
The critic inputs the same denoised observation as the actor, plus some privileged information like the base linear velocity and angular accelerations $\mathbf{v}_b, \mathbf{a}_b \in \mathbb{R}^3$, joint velocities and torques $\dot{\mathbf{q}}, \mathbf{\tau} \in \mathbb{R}^{12}$, the contact state and cartesian velocity $\mathbf{c}_{i}, \mathbf{v}_{c,i} \in \mathbb{R}^{4\cdot3}$, and the force applied to the base $\mathbf{F}_b \in \mathbb{R}^3$ to teach the robot to react to random external disturbances.
\par The policy generates joint position references for the lower-body joints only $\mathbf{\hat{q}} \in \mathbb{R}^{12}$, and sent to the middleware \emph{XBot2}~\cite{laurenzi2023xbot2}, which forward joint references and actuator gains both to simulation environment, like Mujoco, and the real robot.
This allows the sim-to-sim and sim-to-real validation without changing any part of the inference code.
In both cases, joint position references are used to compute joint impedance torques:
\begin{equation}
    \tau = \mathbf{K}(\mathbf{\hat{q}} - \mathbf{q}) - \mathbf{D}\dot{\mathbf{q}}
\end{equation}
These torques are then forwarded as a reference for each actuator.
\section{EXPERIMENTS}
\label{sec:experiments}
In this section, we present a series of experiments aimed at evaluating the robot’s performance and validating the design decisions made during the codesign phase.
The locomotion and manipulation controllers are implemented as two independent modules.
The locomotion controller employs a whole-body Model Predictive Controller (MPC), which extends the Trajectory Optimization (TO) framework used during the design phase through a receding horizon formulation.
Alternatively, the trained policy described in Sec.\ref{sec:control} can be inferenced.
In contrast, the manipulation controller relies on whole-body inverse kinematics (IK) to control and teleoperate one or both arms simultaneously.
Additionally, a simplified omni-steering controller has been implemented for basic wheeled-navigation. This module operates on the two degrees of freedom of each wheel, converting Cartesian velocity references for the base into joint velocities while keeping the rest of the body stationary.
\subsection{Steering wheels}
\label{subsec:steering}
To evaluate the advantage of incorporating steering joints, we compared the cost of transport of the robot while following a predefined path, both with and without steering enabled.
In the first scenario, the robot achieves holonomic motion using all eight wheel joints, whereas in the second scenario it must perform additional maneuvers—such as adjusting its heading or stepping sideways—to change direction.
The configuration with rolling joints only uses a lower leg equipped with a single distal motor for wheel actuation, resulting in a weight reduction of approximately \qty{1.5}{\kilo\gram} compared to the 2-DoF wheel module.
The comparison is conducted by driving the robot along a predefined trajectory that requires omnidirectional motion.
\par The robot with steering joints employs a simple controller that maps a Cartesian velocity reference into joint-space commands for both steering and wheel joints.
Given a velocity command $\hat{\mathbf{v}} = [\mathbf{v}_{\mathrm {cmd}}^T, \mathbf{\omega}_{\mathrm{cmd}}^T]^T = [v_x, v_y, 0, 0, 0, \omega_z]^T$, the reference velocity for each wheel is computed as
$\mathbf{v}_{\mathrm{wheel}, i} = \mathbf{v}_{\mathrm{cmd}} + \omega_{\mathrm{cmd}}\times\mathbf{p}_i$,
where $\mathbf{p}_i$ denotes the vector from the base frame to the $i$-th wheel frame.
The rolling velocity and steering angle for each wheel are then given by
\begin{align}
\mathbf{\dot{q}}_{\mathrm{roll}, i} &= \frac{\mathbf{v}_{\mathrm{wheel}, i}}{r} \\
\mathbf{q}_{\mathrm{steering}, i} &= \mathrm{atan2}(\mathbf{v}_{\mathrm{wheel}, iy}, \mathbf{v}_{\mathrm{wheel}, ix})
\end{align}
The wheeled robot without steering joints, instead, uses a policy similar to the one described in previous section, with an action space enlarged with four extra velocity terms for the wheel continuous joints. 
In this way, the policy learns to use the wheel joints for forward and backward motions, while involving the legs' joints for steering and lateral motion.
\par
\begin{figure}
    \centering
    \includegraphics[width=\columnwidth]{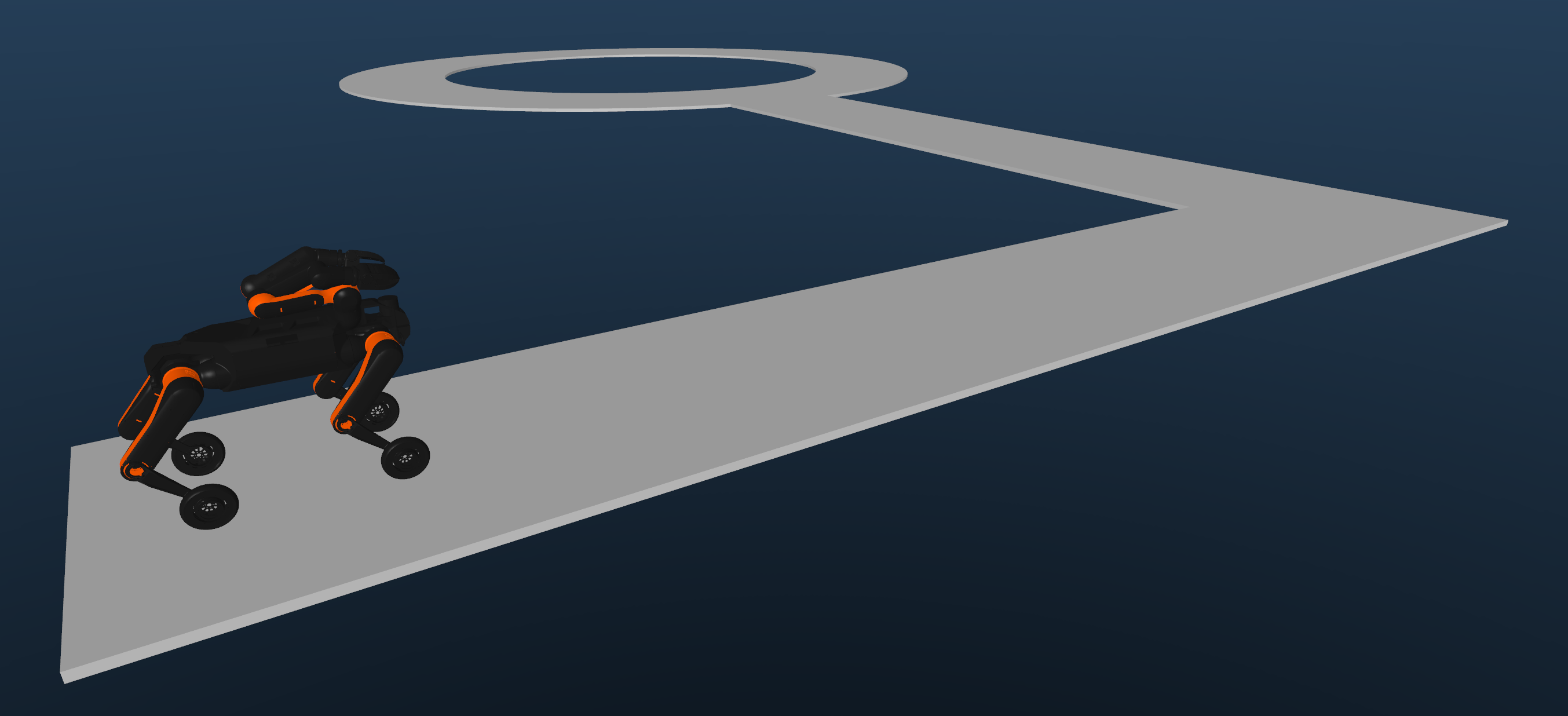}
    \caption{The path used to compare the wheeled platform with and without the steering joints.}
    \label{fig:path}
\end{figure}

 To quantify the benefits of steering, the robot is commanded to follow the path illustrated in Fig.~\ref{fig:path}.
The trajectory consists of a \qty{10}{\meter} straight segment, followed by a \qty{90}{\degree} turn, and then another \qty{10}{\meter} straight segment leading to a circular path with a radius of \qty{5}{\meter}.
This setup enables a comparison of the two configurations over trajectories that include both sharp turns and smooth curves.ng the Cost of Transport (CoT) when completing the path in the two ways. 
The CoT is computed using the classic velocity formulation:
\begin{equation}
    COT = \frac{P}{mgv} = \frac{\mathbf{\tau}_{avg} \cdot \dot{\mathbf{q}}_{avg}^T}{mgv}
\end{equation}
with $P$ being the average mechanical power, $m$ the mass of the robot, $g$ the magnitude of the gravity vector, and $v$ the average of the base planar velocity.
Despite the reduced weight of the quadruped without steering joints, the CoT in this configuration is \qty{0.305}, while the robot with steering joints performs the two-ways path with a CoT equal to \qty{0.102}, highlighting the beneficial of mounting steering joints.
\subsection{Locomotion}
\begin{figure}
\centering
\includegraphics[width=\columnwidth]{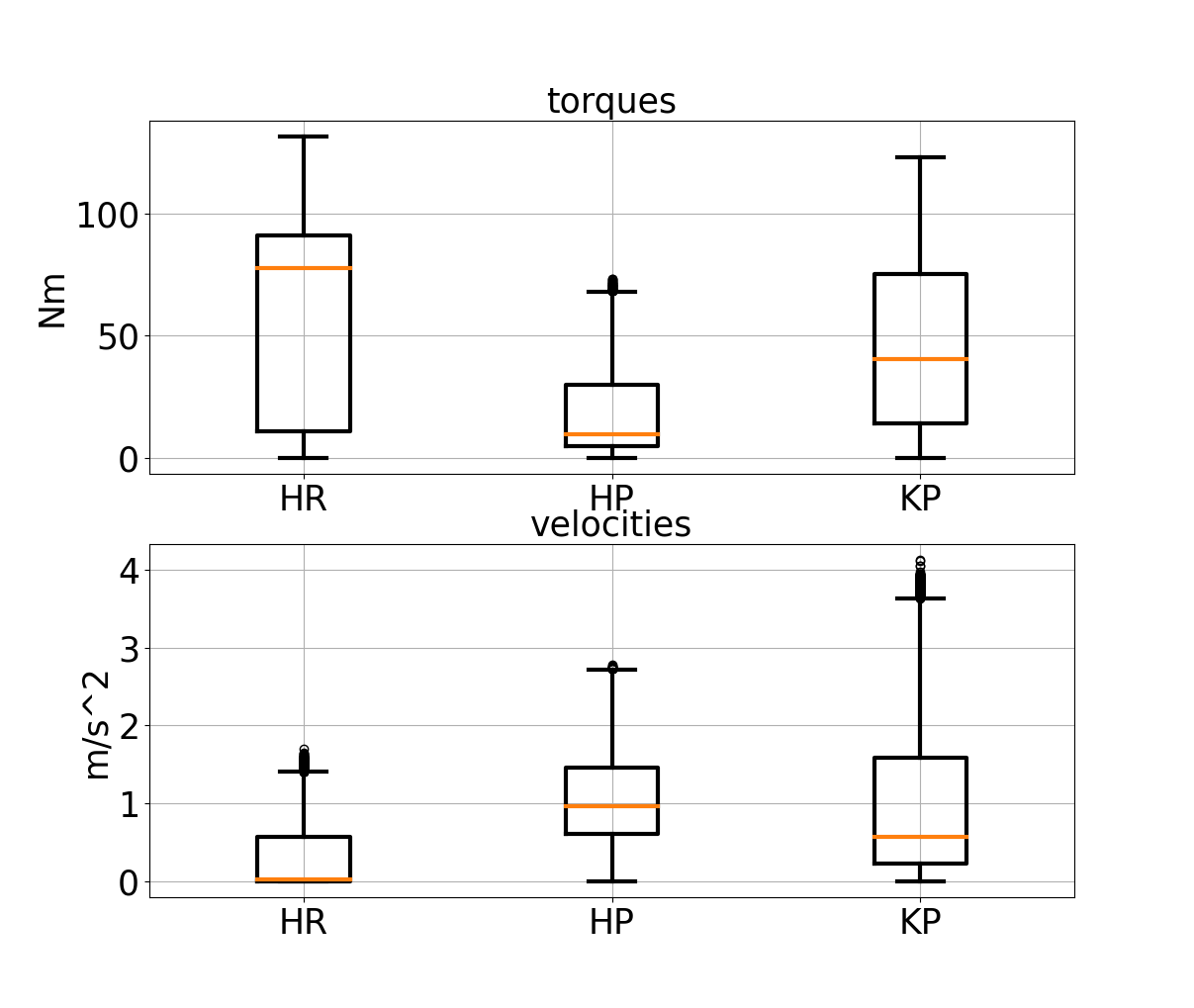}
\caption{Torques and velocities generated during the trot experiment.}
\label{fig:trot_mpc}
\end{figure}

To validate the actuator selection and mechanical design, we ran the robot in its legged configuration using our whole-body MPC and commanded it to trot at a steady speed of \qty{1.0}{\meter\per\second}.
The resulting link-side torques and velocities are shown in Fig.~\ref{fig:trot_mpc}.
Compared to the design-phase simulation results, we observe a consistent trend: the hip-roll joint bears the highest torque loads, while the knee-pitch joint exhibits the fastest movements.
However, the experimental results show a roughly 20\% increase in torque, which can be attributed to variations in motor efficiency across different torque output levels, as well as for the estimated weights of the simplified model which slightly changed during the final design.
\par The zero-shot transfer of the trained policy has been tested both in an indoor and outdoor environment. 
Despite the policy has been trained on flat terrains, it has been tested on the real robot on surfaces with varying slopes and terrains. 
This experiments are intended to validate the trained policy on the real robot only, and consequently its locomotion performance.
They are not shown to assess the performance of the training process, which will be investigated in future works.
In a controlled environment, before bringing the robot outdoor, we used a treadmill with adjustable slopes and speed and run the policy to follow the target speeds. 
Assessing that the robot was able to follow the changes in environment without triggering any safety action from the low-level controller, we decided to bring the robot in a challenging mountain environment.
There, challenges derives not only from the changes in slope, but also from the roughness of the the terrain and by the varying friction conditions of different materials encountered (i.e., stones, mud, and grass).
Videos of the locomotion experiments are in the attached media file.
\subsection{Bi-manual manipulation}
\begin{figure}
\begin{subfigure}{0.45\columnwidth}
\centering
\includegraphics[width=\columnwidth]{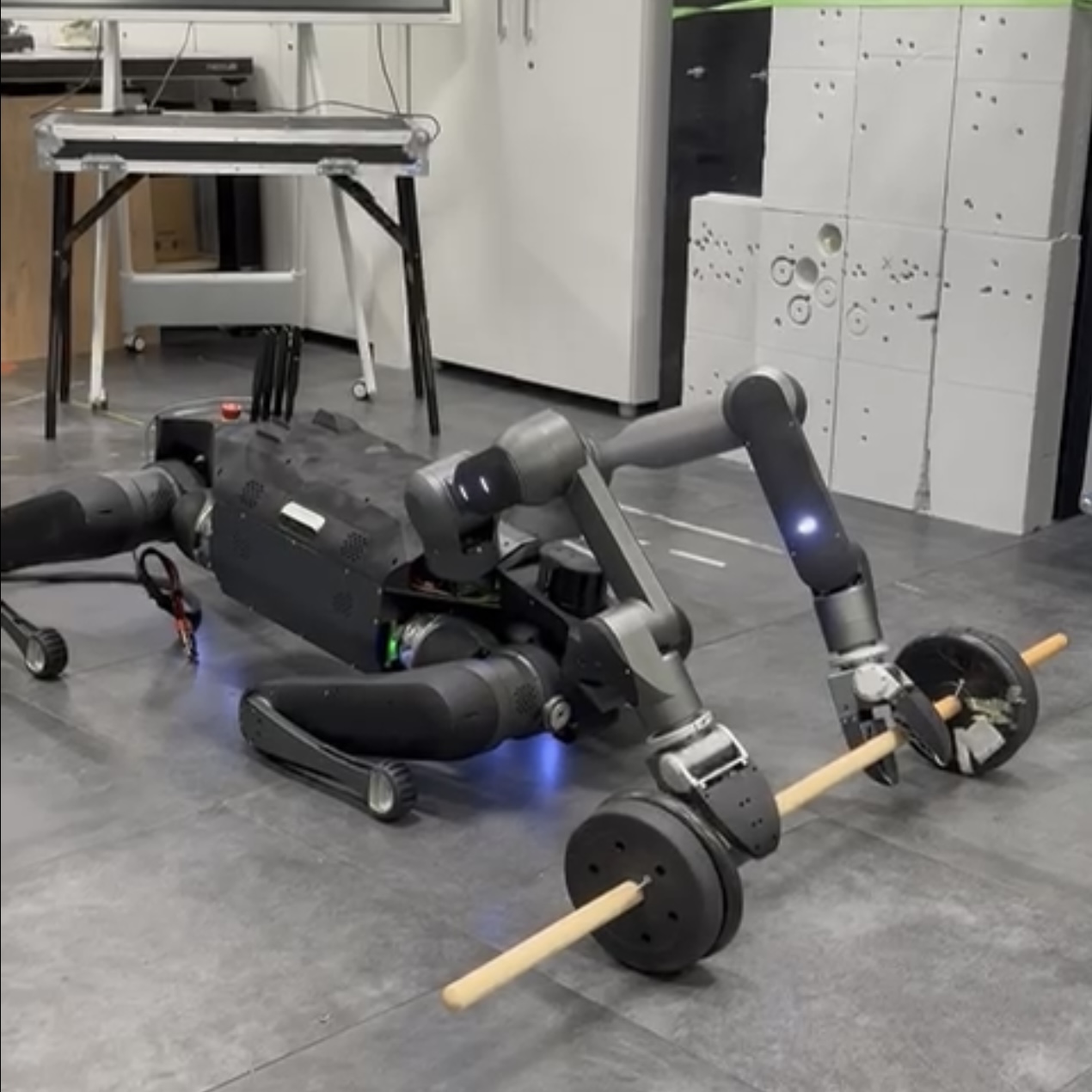}
\end{subfigure}
\quad
\begin{subfigure}{0.45\columnwidth}
\centering
\includegraphics[width=\columnwidth]{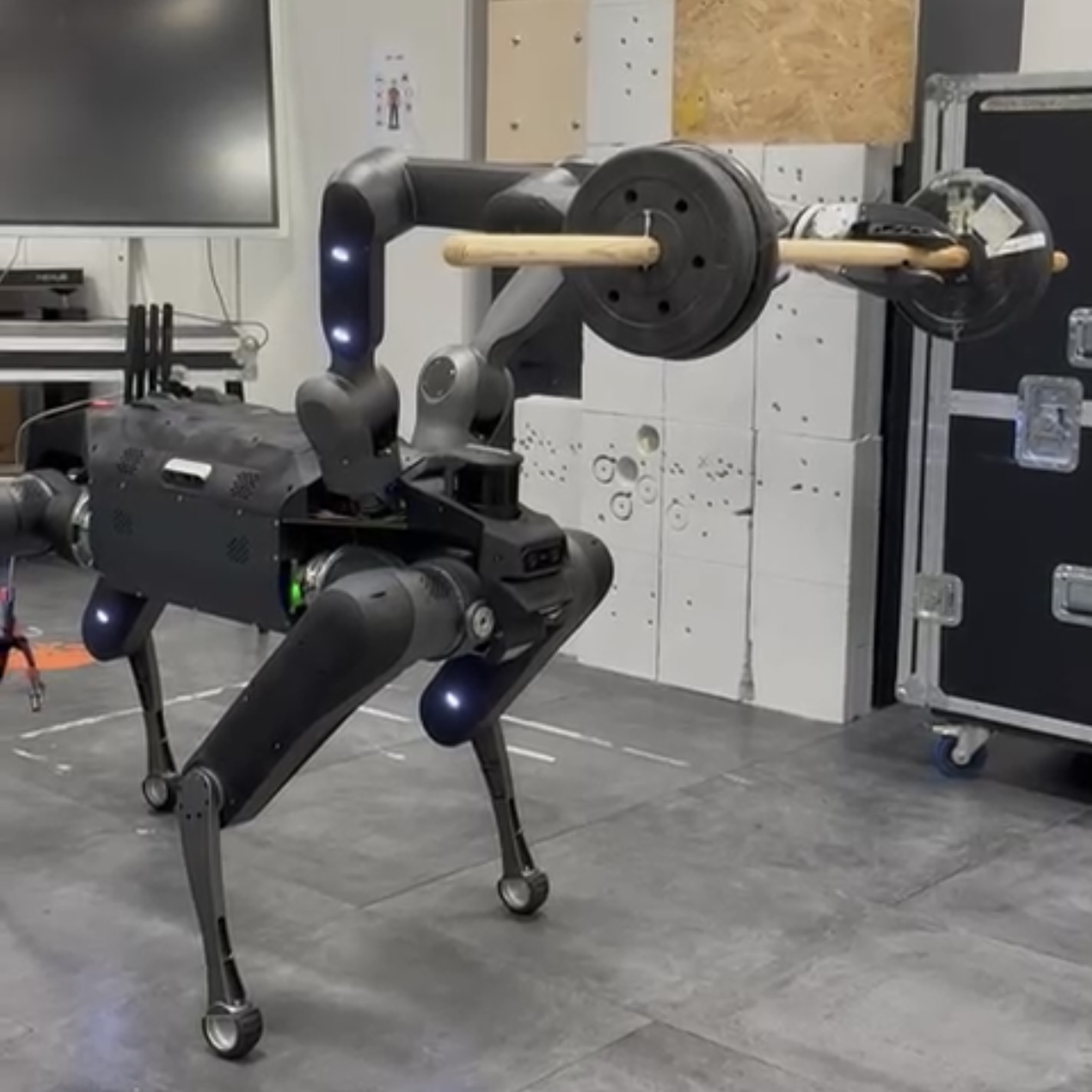}
\end{subfigure}
\caption{Initial and final configuration during the bi-manual grasping experiment lifting \qty{15}{\kilo\gram}}
\label{fig:upper_body_exp}
\end{figure}

The upper-body was evaluated in a demanding task in which the robot was instructed to lift a \qty{15}{\kilo\gram} barbell from the ground to a height of \qty{1.2}{\meter}, reaching a position just above the pelvis. The initial and final poses during the lifting motion are depicted in Fig.~\ref{fig:upper_body_exp}.
\begin{figure}
\centering
\includegraphics[width=\columnwidth]{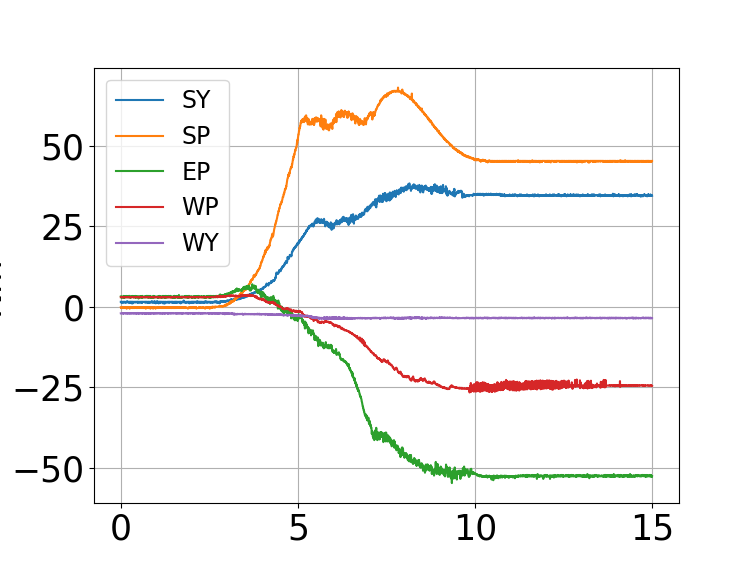}
\caption{Link-side torques generated by the left arm during the bi-manual grasping experiment}
\label{fig:bimanual}
\end{figure}
To assess the system’s capabilities, we measured the joint torques produced by the left arm during the lift, shown in Fig.~\ref{fig:bimanual}.
When compared with the peak torque ratings listed in Tab.~\ref{tab:ub_motors}, the results indicate that the robot has significant torque headroom, suggesting it could handle even heavier loads.
\section{CONCLUSIONS}
\label{sec:conclusions}
In this work, we presented KYON, a novel hybrid wheel–legged quadruped robot equipped with a bi-manual upper body, specifically designed to address high-payload loco-manipulation tasks. The paper detailed the systematic design and co-design methodology adopted to define both the kinematic structure and actuator selection of the lower and upper body, guided by task-driven performance requirements.
Furthermore, the control framework for the execution of locomotion and manipulation tasks is provided.
\par The experimental validation demonstrated the effectiveness of the proposed design choices. In particular, locomotion experiments confirmed the platform’s capability to achieve stable and dynamic motion, while the bi-manual manipulation task highlighted the robot’s ability to handle significant payloads with substantial torque margins. These results indicate that KYON not only satisfies the initial design specifications but also provides additional performance headroom for more demanding applications.
\par Future work will focus on further exploring the limits of the platform, including a comprehensive assessment of maximum achievable locomotion speed and payload capacity. Additionally, more advanced loco-manipulation scenarios will be investigated to fully exploit the potential of the combined wheeled–legged mobility and bi-manual capabilities.
\addtolength{\textheight}{-12cm}   % This command serves to balance the column lengths

\section*{ACKNOWLEDGMENT}
The robot has been developed thanks to the support of the commercial agreement with CETC and SAT.
%%%%%%%%%%%%%%%%%%%%%%%%%%%%%%%%%%%%%%%%%%%%%%%%%%%%%%%%%%%%%%%%%%%%%%%%%%%%%%%%
\bibliographystyle{IEEEtran}
\bibliography{IEEEabrv, bibliography}

\end{document}